\documentclass[numbered,numsec,webpdf,modern,large]{oup-authoring-template}

\usepackage[utf8]{inputenc}
\usepackage[T1]{fontenc}
\usepackage{amsmath,amssymb,amsfonts}
\usepackage{graphicx}
\usepackage{booktabs}
\usepackage{multirow}
\usepackage{float}
\usepackage{microtype}
\usepackage{url}
\usepackage{balance}

\definecolor{linkcolor}{RGB}{0, 51, 102}
\graphicspath{{figures/}}
\newcommand{\pathmog}{\textbf{PathMoG}}

\begin{document}

\journaltitle{}
\copyrightyear{}
\pubyear{}
\appnotes{}
\makeatletter
\let\ps@opening\ps@plain
\makeatother

\title[PathMoG: A Pathway-Centric Modular GNN]{\pathmog: A Pathway-Centric Modular Graph Neural Network for Multi-Omics Survival Prediction}

\author[1]{Di Wang}
\author[2]{Chupei Tang}
\author[3]{Junxiao Kong}
\author[4]{Jixiu Zhai}
\author[5]{Moyu Tang}
\author[6,$\ast$]{Tianchi Lu}

\address[1]{\orgdiv{Cuiying Honors College}, \orgname{Lanzhou University}, \orgaddress{\country{China}}}
\address[2]{\orgdiv{School of Mathematics and Statistics}, \orgname{Lanzhou University}, \orgaddress{\country{China}}}
\address[3]{\orgdiv{School of Mathematics and Statistics}, \orgname{Lanzhou University}, \orgaddress{\country{China}}}
\address[4]{\orgdiv{School of Mathematics and Statistics}, \orgname{Lanzhou University}, \orgaddress{\country{China}} and \orgname{Shanghai Innovation Institute}}
\address[5]{\orgdiv{School of Mathematics and Statistics}, \orgname{Lanzhou University}, \orgaddress{\country{China}}}
\address[6]{\orgdiv{Department of Computer Science}, \orgname{City University of Hong Kong}, \orgaddress{\country{China}}}

\corresp[$\ast$]{Corresponding author. Email: tianchilu4-c@my.cityu.edu.hk}

\abstract{
\textbf{Motivation:} Cancer survival prediction from multi-omics data remains challenging because prognostic signals are high dimensional, heterogeneous, and distributed across interacting genes and pathways. Existing survival models either ignore biological structure, rely on genome-scale monolithic graphs, or fuse expression, mutation, and copy number variation (CNV) too coarsely for the low-sample, high-dimensional setting. \\
\textbf{Results:} In this article, we propose \pathmog{} (Pathway-centric Modular Omics Graph), a biologically structured multi-omics graph neural network for cancer survival prediction. \pathmog{} first reorganizes genome-scale inputs into 354 KEGG-informed pathway modules, introducing functional priors that constrain representation learning in the high-dimensional, low-sample regime. We then use a Hierarchical Omics Modulation (HOM) module to condition gene-expression representations on mutation, CNV, pathway, and clinical context, enabling semantically aligned cross-omics integration rather than coarse feature concatenation. A dual-level attention mechanism further captures both intra-pathway driver signals and inter-pathway clinical relevance to generate patient-level Cox risk estimates with multi-level interpretability. We comprehensively evaluated \pathmog{} on 5,650 patients across 10 TCGA cancer types, and the results demonstrate clear superiority over representative survival baselines, while the framework provided interpretable gene-level, pathway-level, and patient-level readouts for biologically grounded and clinically actionable risk stratification. \\
\textbf{Availability and implementation:} TCGA data were accessed through UCSC Xena and external validation used METABRIC. Source code is available at \url{https://github.com/wangzoyou/pathmog}.
}


\maketitle

\section{Introduction}

Cancer prognosis is intrinsically a multi-factor problem, because clinically similar patients can follow markedly different trajectories driven by distinct molecular programs \cite{Siegel2023, Steele2022}. In modern cohorts, prognostic signals are rarely confined to one modality and instead span expression, mutation, and copy-number variation (CNV), making multi-omics integration essential for risk modeling \cite{Huang2017Multiomics}. At the same time, cancer datasets typically operate under a high-dimensional low-sample regime ($p \gg n$), where robust generalization is difficult.

Methodologically, survival modeling still builds on three classical families: non-parametric, parametric, and semi-parametric formulations \cite{Bradburn2003}. Cox proportional hazards and its extensions remain central in biomedical applications \cite{Cox1972}, and random survival forest provides a strong nonlinear baseline \cite{Ishwaran2008}. These approaches are statistically grounded and interpretable, but they are not designed to directly capture the high-order cross-gene interactions and cross-omics dependencies that dominate modern cancer data.

Deep survival and multimodal fusion models emerged to address this representational bottleneck. CAMR introduced cross-aligned multimodal representation learning to improve cross-omics consistency in prognosis prediction \cite{Jiang2022_CAMR}; HFBSurv used hierarchical fusion with factorized bilinear interactions to model cross-modal feature coupling \cite{Li2022_HFB}; and PCLSurv incorporated prototypical contrastive learning to strengthen class-aware multi-omics representations for survival modeling \cite{Li2025_PCL}. However, many of these methods still rely on coarse fusion, where omics views are treated as parallel feature blocks and merged by concatenation or weakly constrained attention, which can blur biological directionality between upstream genomic alterations and downstream transcriptional readouts.

Graph-based prognosis models were introduced as a second-stage correction to this limitation by injecting prior biological structure. Representative studies follow a clear progression: GraphSurv combined GCN encoding with a Cox survival objective for multi-omics prognosis \cite{Wang2021_BIBM}; PathGNN introduced pathway-oriented interpretable graph modeling for risk stratification and pathway analysis \cite{Liang2022_BMC}; FGCNSurv jointly learned omics feature representations and feature-relation structure for pan-cancer survival prediction \cite{Wen2023_FGCN}; and prior knowledge-guided multilevel GNNs further strengthened biologically constrained prognosis modeling \cite{Yan2024}. In adjacent but highly relevant graph-centric settings, DeepKEGG used biologically informed graph modeling to capture cell heterogeneity from single-cell and spatial transcriptomics data, while PathHDNN used pathway-hierarchical deep architecture for immunotherapy response prediction and mechanism interpretation \cite{Lan2024, Li2025_PathHDNN}. In adjacent tasks such as multi-omics cancer subtyping, graph contrastive learning has also reinforced the value of structure-aware representations \cite{Yang2025M2CGCN}. Yet many graph pipelines remain monolithic at genome scale, and this ``one-big-graph'' strategy often introduces three practical issues in $p \gg n$ cohorts: noise propagation through task-irrelevant long-range edges, explanatory dilution of patient-specific drivers, and excessive hypothesis complexity that increases overfitting risk.

Taken together, the field has progressed from black-box fusion toward biologically informed graphs, but a key tension remains: prior biology is necessary, while unconstrained global graphs are often too blunt for stable survival learning. This tension defines two gaps addressed in this work: a \textit{graph structure gap} (how to encode biology as a useful constraint rather than a massive graph prior) and an \textit{omics semantics gap} (how to preserve directional relations across omics layers rather than treating them as symmetric channels).

\begin{figure*}[t]
\centering
\includegraphics[width=\textwidth]{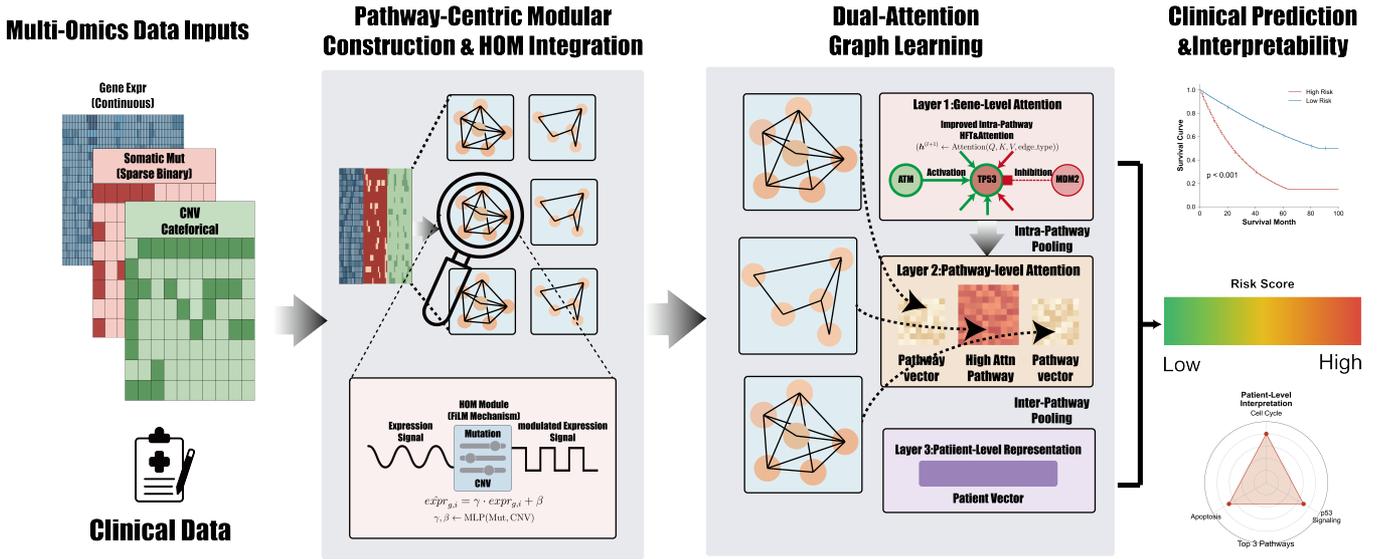}
\caption{\textbf{Workflow overview of \pathmog{}.} The framework combines pathway-centric graph construction, HOM-based omics modulation, heterogeneous message passing within each pathway module, dual-level attention, and Cox-based survival prediction with clinical fusion.}
\label{fig:method_flow}
\end{figure*}

To resolve these gaps, we propose \pathmog{}, a pathway-centric modular graph neural network for multi-omics survival prediction. Instead of a monolithic graph, \pathmog{} decomposes the interactome into 354 KEGG-informed predefined pathway modules; applies a Hierarchical Omics Modulation (HOM) mechanism to condition expression on mutation, CNV, pathway, and clinical context via FiLM-style modulation; and aggregates evidence through dual-level attention for patient-level Cox risk estimation with interpretable outputs. We evaluate \pathmog{} across 10 TCGA cancer types and further examine transferability on METABRIC, clinical relevance in BRCA, and mechanistic validity through ablation and interpretability analyses. The end-to-end workflow is shown in Figure~\ref{fig:method_flow}.

\section{Materials and methods}

\subsection{Datasets}

We evaluated \pathmog{} on a harmonized pan-cancer cohort of 5,650 patients from 10 TCGA cancer types, where each patient was represented by gene expression, somatic mutation, CNV, and survival outcome information. These cohorts span diverse tissue origins, event rates, and sample sizes, making them suitable for assessing generalizability across heterogeneous disease settings. We additionally reserved the METABRIC breast cancer cohort as an external transfer cohort for independent testing without model fine-tuning. Cohort-wise sample sizes and event rates are summarized in Table~\ref{tab:dataset_info}.
\begin{table}[h]
\centering
\caption{\textbf{Dataset characteristics for 10 TCGA cancer types.} The cohort covers diverse tissue origins and survival event rates, providing a rigorous testbed for generalizability. All cancer types use 7,595 genes from 354 KEGG pathways, and the total event rate is reported as a sample-size-weighted average across cohorts.}
\label{tab:dataset_info}
\resizebox{\linewidth}{!}{
\begin{tabular}{@{}lrr@{}}
\toprule
\textbf{Cancer Type} & \textbf{Samples} & \textbf{Event Rate (\%)} \\ 
\midrule
BRCA (Breast Invasive Carcinoma) & 1,082 & 14.0 \\ 
COAD (Colon Adenocarcinoma) & 430 & 22.2 \\ 
GBM (Glioblastoma Multiforme) & 580 & 80.6 \\ 
KIRC (Kidney Renal Clear Cell) & 534 & 33.1 \\ 
LGG (Lower Grade Glioma) & 511 & 24.5 \\ 
LIHC (Liver Hepatocellular) & 371 & 35.6 \\ 
LUAD (Lung Adenocarcinoma) & 506 & 36.1 \\ 
UCEC (Uterine Corpus Endometrial) & 541 & 16.5 \\ 
HNSC (Head \& Neck Squamous) & 525 & 35.1 \\ 
OV (Ovarian Serous Cystadenocarcinoma) & 570 & 59.7 \\ 
\midrule
\textbf{Total} & \textbf{5,650} & \textbf{34.4} \\ 
\bottomrule
\end{tabular}
}
\end{table}

Expression profiles were log-transformed and standardized, somatic mutation calls were binarized at the gene level, and GISTIC CNV calls were encoded on a five-level scale from $-2$ to $+2$ \cite{hoadley2018cell, Mermel2011_GISTIC}. Let $\mathcal{I}$ denote the patient set used for modeling. In the implementation, $\mathcal{I}$ is defined by the intersection of patients with both clinical and survival annotations, after which all omics matrices are reindexed to a common patient order. Likewise, let $\mathcal{U}=\{g_1,\dots,g_M\}$ denote the aligned gene universe, where $M=7{,}595$ genes are induced by the curated gene--pathway mapping used throughout the study. Expression, CNV, and mutation matrices are reindexed to this common gene order before pathway-level graph instantiation. Missing modality-specific entries are preserved during alignment and converted to zero only at the final tensor-materialization stage, which avoids mixing patient alignment with imputation. Clinical covariates were compressed into a numerically stable feature vector containing age, sex, and aggregated stage variables, with cancer-specific markers retained when clinically standard information was available, such as receptor status and PAM50 subtype in BRCA. All preprocessing parameters were estimated within training folds to avoid information leakage. Detailed preprocessing, cohort descriptors, and leakage-prevention rules are provided in Supplementary Sections S1, S3, and S4.

\subsection{Overview of \pathmog{}}

\pathmog{} follows a pathway-first pipeline. Multi-omics inputs are first mapped onto KEGG pathway graphs, then gene expression is modulated by genomic and clinical context through HOM, pathway-specific node embeddings are updated with heterogeneous graph message passing, pathway vectors are aggregated through gene-to-pathway and pathway-to-patient attention, and the final patient representation is fused with clinical covariates for Cox-based survival prediction. A key implementation feature is that the model does not operate on a single fixed graph tensor. Instead, it uses a nested representation in which pathway topologies are precomputed once and each patient is instantiated as a variable-length collection of pathway graphs plus one clinical vector. This hierarchical data structure is central to both the computational tractability and the biological interpretability of the framework.

\subsection{Pathway-centric modular graph construction}

The central design choice in \pathmog{} is to replace a single global gene graph with pathway modules. This modular design introduces an explicit biological prior, constrains the search space, and reduces spurious interactions in the $p \gg n$ regime, rather than asking the model to infer all genome-wide dependencies from limited patient samples \cite{Hartwell1999, kanehisa2000kegg, kanehisa2023kegg}. In this sense, pathway modularity acts as an inductive bias for survival modeling rather than as a visualization convenience.

Formally, let $\mathcal{P}=\{1,\dots,K\}$ denote the pathway catalogue with $K=354$ modules. The implementation first constructs a patient-invariant library of local pathway topologies. For pathway $k\in\mathcal{P}$, the static graph is
$$
\mathcal{G}^{\mathrm{topo}}_k=\big(\mathcal{V}_k,\{\mathcal{E}^{(r)}_k\}_{r\in\mathcal{R}_k}\big),
$$
where $\mathcal{V}_k\subseteq\mathcal{U}$ is the set of genes belonging to pathway $k$ after intersection with the global aligned gene universe, and
$$
\mathcal{E}^{(r)}_k=\{(u,v):u,v\in\mathcal{V}_k,\ \mathrm{rel}(u,v)=r\}
$$
collects all directed edges of relation type $r$ whose two endpoints both remain inside the pathway. Relation labels are normalized from the curated pathway interaction file and stored explicitly as typed edges in \texttt{HeteroData}. Pathways retaining fewer than two mapped genes after alignment are not instantiated. This static precomputation step fixes a pathway-local node ordering and avoids rebuilding the same biological topology for every patient.

For patient $i\in\mathcal{I}$ and pathway $k$, omics measurements are then injected into the precomputed topology through a pathway-local feature matrix
$$
\mathbf{X}_{k,i}\in\mathbb{R}^{|\mathcal{V}_k|\times 3},\qquad
\mathbf{X}_{k,i}(u,:)=[x^{\mathrm{expr}}_{u,i},\ x^{\mathrm{cnv}}_{u,i},\ x^{\mathrm{mut}}_{u,i}],
$$
where the column order follows the implementation exactly: expression, CNV, then mutation. If all three modalities are absent for every gene in pathway $k$ for patient $i$, that pathway is omitted from the patient's package; otherwise missing entries are mapped to $0$ only when the tensor is materialized. The resulting patient-level input is therefore not a monolithic graph but a hierarchical object
$$
\mathcal{D}_i=\big(\{G_{k,i}\}_{k\in\mathcal{K}_i},\ \mathbf{c}_i\big),
$$
where $\mathcal{K}_i\subseteq\mathcal{P}$ is the set of pathway graphs instantiated for patient $i$, $G_{k,i}$ carries the typed pathway edges together with node features $\mathbf{X}_{k,i}$, and $\mathbf{c}_i$ is the processed clinical feature vector. Supplementary Section S4 provides the full graph-construction rules, patient-package definition, and code-level batching details.

\subsection{Hierarchical Omics Modulation (HOM)}

Rather than treating multi-omics inputs as interchangeable channels, we model gene expression as the target signal modulated by genomic and clinical context. This reflects the biological intuition that mutation and CNV states act as regulatory conditions that shape the functional meaning of transcript abundance, while pathway identity and patient context provide higher-order information about the state in which a gene is being interpreted.

In the implemented model, HOM conditions each gene on three nested contexts. For gene $u\in\mathcal{V}_k$ in patient $i$, define the local genomic context
$$
\mathbf{m}_{u,k,i}=[x^{\mathrm{cnv}}_{u,i},\ x^{\mathrm{mut}}_{u,i}]\in\mathbb{R}^2,
$$
the encoded patient context
$$
\mathbf{z}_i=f_{\mathrm{clin}}(\mathbf{c}_i),
$$
and the pathway context
$$
\bar{\mathbf{m}}_{k,i}=\frac{1}{|\mathcal{V}_k|}\sum_{u\in\mathcal{V}_k}\mathbf{m}_{u,k,i},
\qquad
\mathbf{p}_{k,i}=f_{\mathrm{path}}([\mathbf{e}_k;\bar{\mathbf{m}}_{k,i}]),
$$
where $\mathbf{e}_k$ is a learnable embedding representing pathway identity. HOM then forms the context vector
$$
\mathbf{ctx}_{u,k,i}=[\mathbf{m}_{u,k,i};\mathbf{p}_{k,i};\mathbf{z}_i]
$$
and feeds it to a modulation network. Following a FiLM-style parameterization \cite{Perez2018_FiLM}, the network predicts raw coefficients $(\hat\gamma_{u,k,i},\hat\beta_{u,k,i})$, which are range-controlled in the implementation as
$$
\gamma_{u,k,i}=1+0.5\tanh(\hat\gamma_{u,k,i}),
\qquad
\beta_{u,k,i}=2\tanh(\hat\beta_{u,k,i}),
$$
so that the modulated expression remains numerically stable while still being patient- and pathway-specific. The actual modulation is
$$
\tilde{x}^{\mathrm{expr}}_{u,k,i}=\gamma_{u,k,i}\,x^{\mathrm{expr}}_{u,i}+\beta_{u,k,i}.
$$
The initial node representation passed to the graph encoder is then the fused vector
$$
\mathbf{h}^{(0)}_{u,k,i}=
[\tilde{x}^{\mathrm{expr}}_{u,k,i},\ x^{\mathrm{expr}}_{u,i},\ x^{\mathrm{cnv}}_{u,i},\ x^{\mathrm{mut}}_{u,i},\ \gamma_{u,k,i},\ \beta_{u,k,i},\ \mathbf{e}_k,\ \bar{\mathbf{m}}_{k,i}],
$$
which has dimension $16$ in the current implementation. Hence, HOM does not merely rescale one channel; it creates an explicit, inspectable representation in which modulation coefficients, pathway identity, and pathway state all remain accessible to downstream analysis. Full mathematical details are provided in Supplementary Section S4.2.

\subsection{Heterogeneous graph learning and dual-level attention}

Within each pathway module, \pathmog{} applies Heterogeneous Graph Transformer (HGT) layers to account for typed regulatory edges such as activation, inhibition, and phosphorylation \cite{Hu2020_HGT}. Each instantiated pathway graph contains one node type (gene) and multiple relation types inherited from the pathway interaction file. If $\mathbf{h}^{(\ell)}_v$ denotes the embedding of gene $v$ at layer $\ell$, then the relation-aware update can be written abstractly as
$$
\mathbf{h}_v^{(\ell+1)}=
\sigma\!\left(
\sum_{r\in\mathcal{R}_k}
\sum_{u\in\mathcal{N}^{(r)}(v)}
\alpha_{uv}^{(\ell,r)}\,\mathbf{W}^{(\ell)}_r\mathbf{h}^{(\ell)}_u
\right),
$$
where $\mathcal{N}^{(r)}(v)$ denotes the neighbors of $v$ connected by relation type $r$, and the attention coefficients $\alpha_{uv}^{(\ell,r)}$ depend jointly on source state, target state, and relation type. In the current implementation, two HGT layers are used inside each pathway module.

After message passing, node embeddings are summarized through two attention stages. For pathway $k$ in patient $i$, gene-to-pathway attention computes
\[
\begin{aligned}
a_{u,k,i}
&=\mathbf{w}_g^\top\tanh(\mathbf{W}_g\mathbf{h}^{(L)}_{u,k,i}),\\
\alpha_{u,k,i}
&=\frac{\exp(a_{u,k,i})}{\sum_{v\in\mathcal{V}_k}\exp(a_{v,k,i})},
\end{aligned}
\]
and forms the pathway vector
$$
\mathbf{g}_{k,i}=\sum_{u\in\mathcal{V}_k}\alpha_{u,k,i}\mathbf{h}^{(L)}_{u,k,i}.
$$
Pathway-to-patient attention then aggregates the variable-length pathway set $\{\mathbf{g}_{k,i}\}_{k\in\mathcal{K}_i}$ by
\[
\begin{aligned}
b_{k,i}
&=\mathbf{w}_p^\top\tanh(\mathbf{W}_p\mathbf{g}_{k,i}),\\
\pi_{k,i}
&=\frac{\exp(b_{k,i})}{\sum_{\ell\in\mathcal{K}_i}\exp(b_{\ell,i})},\\
\mathbf{H}_i
&=\sum_{k\in\mathcal{K}_i}\pi_{k,i}\mathbf{g}_{k,i}.
\end{aligned}
\]
This representation is fused with encoded clinical features and passed to a multilayer perceptron to produce the final survival risk score.

The corresponding batching strategy is also hierarchical. A custom collate function concatenates all pathway graphs from all patients in a minibatch into a single PyG batch, while retaining three mappings: the gene-to-pathway batch index generated automatically by PyG, an explicit pathway-to-patient map, and the global pathway index vector used to retrieve pathway embeddings. These mappings allow patient-level clinical context to be broadcast to pathways and then to genes without padding every patient to a fixed number of pathways, which is essential because different patients instantiate different subsets of pathway modules after modality-aware filtering.

\begin{table*}[t]
\centering
\caption{\textbf{Concordance Index (C-index) and Time-dependent AUC across 10 TCGA cancer types.} Per-cohort entries are mean test-set metrics from 5-fold cross-validation, and the Overall row reports sample-size-weighted averages across the 10 cohorts. \pathmog{} ranks first in C-index for all 10 cohorts and in AUC for 8 of 10 cohorts.}
\label{tab:overall_performance}
\begingroup
\scriptsize
\setlength{\tabcolsep}{2.8pt}
\renewcommand{\arraystretch}{1.10}
\resizebox{\textwidth}{!}{%
\begin{tabular}{@{}lrrrrrrrrrr@{}}
\toprule
\multirow{2}{*}{\textbf{Cancer}} & \multirow{2}{*}{\textbf{N}} & \multicolumn{9}{c}{\textbf{C-index/AUC}} \\
\cmidrule(lr){3-11}
 & & \textbf{Cox} & \textbf{RSF} & \textbf{CAMR} & \textbf{PathG} & \textbf{GraphS} & \textbf{FGCN} & \textbf{HFB} & \textbf{PCL} & \textbf{\pathmog{}} \\
\midrule
BRCA & 1082 & 0.712/0.663 & 0.695/0.642 & 0.667/0.717 & 0.685/0.776 & 0.640/0.687 & 0.756/0.690 & 0.671/0.627 & 0.760/0.740 & \textbf{0.784/0.811}  \\
KIRC & 534 & 0.726/0.626 & 0.718/0.619 & 0.702/0.756 & 0.658/0.675 & 0.687/0.652 & 0.742/0.738 & 0.574/0.581 & 0.748/0.765 & \textbf{0.751/0.780}  \\
GBM & 580 & 0.520/0.550 & 0.542/0.573 & 0.659/0.650 & 0.569/0.605 & 0.553/0.592 & 0.610/0.592 & 0.563/0.443 & 0.654/0.665 & \textbf{0.724/0.810}  \\
LUAD & 506 & 0.607/0.547 & 0.615/0.560 & 0.651/0.676 & 0.605/0.660 & 0.555/0.618 & 0.605/0.702 & 0.584/0.563 & 0.683/0.630 & \textbf{0.709/0.757}  \\
COAD & 430 & 0.517/0.546 & 0.538/0.567 & 0.650/0.660 & 0.597/0.667 & 0.504/0.623 & 0.650/0.700 & 0.673/0.477 & 0.650/0.680 & \textbf{0.694/0.699}  \\
LGG & 511 & 0.713/0.738 & 0.702/0.727 & 0.704/0.732 & 0.692/0.718 & 0.711/0.732 & 0.720/0.725 & 0.698/0.735 & 0.715/0.748 & \textbf{0.753/0.814}  \\
HNSC & 525 & 0.590/0.673 & 0.582/0.655 & 0.595/0.672 & 0.600/0.658 & 0.585/0.672 & 0.605/0.660 & 0.588/0.638 & 0.602/0.650 & \textbf{0.610/0.668}  \\
OV & 570 & 0.564/0.617 & 0.572/0.625 & 0.645/0.671 & 0.576/0.637 & 0.617/0.671 & 0.609/0.598 & 0.594/0.548 & 0.600/0.680 & \textbf{0.629/0.691}  \\
UCEC & 541 & 0.503/0.531 & 0.519/0.548 & 0.592/0.624 & 0.568/0.585 & 0.503/0.602 & 0.598/0.602 & 0.582/0.532 & 0.600/0.623 & \textbf{0.680/0.714}  \\
LIHC & 371 & 0.626/0.598 & 0.642/0.615 & 0.625/0.661 & 0.647/0.653 & 0.597/0.638 & 0.652/0.580 & 0.524/0.563 & 0.640/0.650 & \textbf{0.654/0.666}  \\
\midrule
\textbf{Overall (weighted)} & \textbf{5650} & 0.618/0.615 & 0.620/0.616 & 0.651/0.686 & 0.625/0.674 & 0.601/0.653 & 0.664/0.662 & 0.612/0.576 & 0.675/0.689 & \textbf{0.708/0.751}  \\
\bottomrule
\end{tabular}%
}
\endgroup

\begin{flushleft}
\footnotesize Cox: Cox-PH; RSF: random survival forest; CAMR: cross-modal attention; PathG: PathGNN (Liang et al., BMC Bioinformatics 2022); GraphS: GraphSurv (Wang et al., BIBM 2021); FGCN: FGCNSurv; HFB: HFBSurv; PCL: PCLSurv. AUC: Area Under ROC Curve at median survival time. \pathmog{} achieves the highest C-index in all 10 cancer types.
\end{flushleft}
\end{table*}
\subsection{Survival objective and evaluation protocol}

Model parameters were optimized with Cox partial likelihood and $L_2$ regularization,
$$\mathcal{L}(\theta) = -\sum_{i:E_i=1}\left(\theta_i - \log\sum_{j\in\mathcal{R}(t_i)} e^{\theta_j}\right)+\lambda\lVert\mathbf{W}\rVert_2^2,$$
where $\theta_i$ is the predicted risk score for patient $i$, $E_i$ is the event indicator, and $\mathcal{R}(t_i)$ is the risk set at time $t_i$ \cite{Bradburn2003}.

We used stratified 5-fold cross-validation within each TCGA cohort. The concordance index (C-index) served as the primary metric, with time-dependent AUC used as a secondary metric. Detailed implementation, evaluation metrics, and reproducibility settings are provided in Supplementary Sections S2--S4.

\section{Results}

\subsection{Benchmark performance}

We benchmarked \pathmog{} against eight representative methods under the same stratified 5-fold protocol: Cox-PH, RSF, GraphSurv (Wang et al., BIBM 2021), CAMR, PathGNN (Liang et al., BMC Bioinformatics 2022), HFBSurv, FGCNSurv, and PCLSurv. This comparison spans classical survival analysis, deep survival learning, multi-omics integration, and graph-based survival modeling, providing a broad view of how \pathmog{} behaves relative to established baselines.

All methods were trained under a consistent evaluation protocol (Supplementary Sections S2--S4). Under this controlled benchmark, \pathmog{} delivered the best overall discrimination performance across TCGA cohorts (Table~\ref{tab:overall_performance}), ranking first in C-index in all 10 cancer types and showing especially strong gains in GBM, UCEC, BRCA, and COAD. These results provide direct evidence that pathway-constrained modular learning is more effective than unconstrained global interaction modeling when prognostic signals are noisy, heterogeneous, and distributed across biological programs.

The superiority of \pathmog{} also transferred to external testing: a model trained on TCGA BRCA maintained significant risk separation in METABRIC without fine-tuning, indicating that the learned representations generalize beyond a single cohort distribution. Across TCGA cohorts, the \pathmog{} score likewise achieved significant high-versus-low risk stratification in Kaplan--Meier analysis (all log-rank $p<0.05$), supporting clinical utility beyond point-estimate ranking metrics. Full external-validation results and cohort-wise Kaplan--Meier curves are provided in Supplementary Sections S5 and S6, respectively. Having established the overall performance advantage of \pathmog{}, we next examined which architectural choices produced that gain.

\subsection{Architecture validation}

To explain why \pathmog{} performs well, we next tested whether pathway modularity acts as a genuine regularizer rather than a cosmetic architectural choice.

\subsubsection{Modular versus monolithic graph design}

We first compared \pathmog{} with a monolithic graph model built on the same 7,595 pathway genes but merged into a single topology. Under this matched comparison, the difference between the two designs lies not in the gene set itself but in the structural constraints imposed on learning.

In the monolithic model, all genes are embedded within a single global interaction network, allowing message passing to propagate across distant and potentially unrelated interactions. Because each patient then has only one graph instance, the pathway-level aggregation stage is removed in this baseline and the single graph representation is fed directly to the patient-level survival head. In contrast, \pathmog{} organizes the same genes into biologically curated pathway modules, restricting information propagation to functionally coherent neighborhoods defined by pathway topology.

This modular organization introduces a biologically informed inductive bias that constrains the hypothesis space of the model. Rather than exploring genome-scale interactions indiscriminately, \pathmog{} limits representation learning to pathway-structured contexts, which helps suppress spurious correlations that frequently arise in high-dimensional survival modeling ($p \gg n$).

\begin{figure*}[t]
\centering
\includegraphics[width=\textwidth]{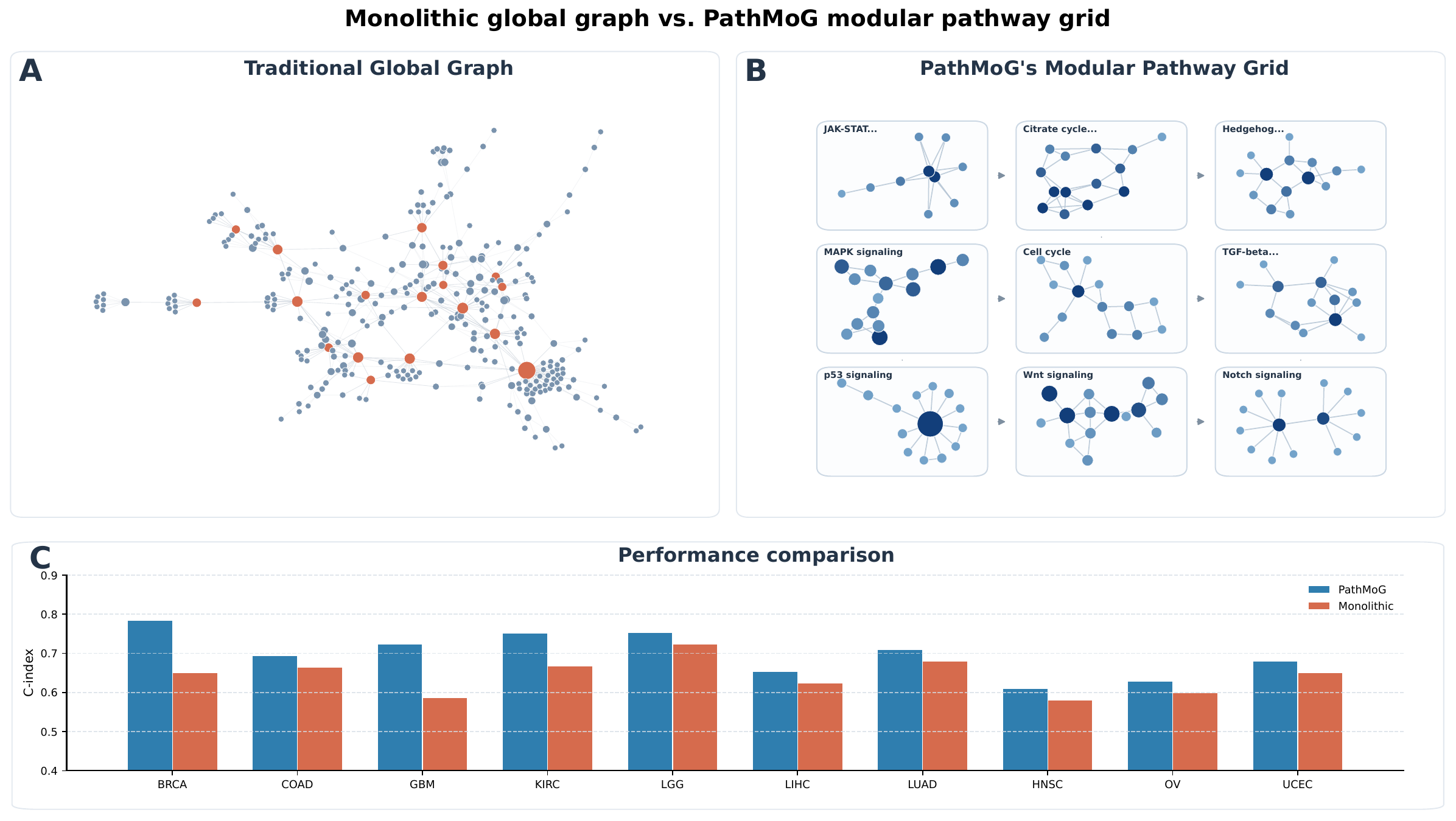}
\caption{\textbf{Comparison of monolithic and modular graph designs.} \textbf{(A)} Traditional global graph architecture with all 7,595 pathway genes merged into a single topology. \textbf{(B)} PathMoG's modular pathway grid organizing the same genes into biologically curated pathway modules. \textbf{(C)} Performance comparison showing PathMoG consistently outperforms the monolithic baseline across all 10 TCGA cancer types.}
\label{fig:pathway_comparison}
\end{figure*}

Consistent with this hypothesis, the monolithic baseline underperformed overall on a sample-size-weighted basis (C-index 0.643 vs.~0.708; +10.1\% relative gain for \pathmog{}; $p < 0.001$), with the largest performance gaps observed in GBM and BRCA (Figure~\ref{fig:pathway_comparison}C). Detailed cohort-wise comparisons and the exact monolithic-baseline construction (including removal of pathway-level aggregation) are provided in Supplementary Section~S8.

Taken together, these results suggest that the advantage of \pathmog{} does not arise from reducing model size, but from replacing an unconstrained genome-scale interaction space with a biologically structured learning framework.

\subsubsection{Component ablation}

\begin{figure}[t]
\centering
\includegraphics[width=\columnwidth]{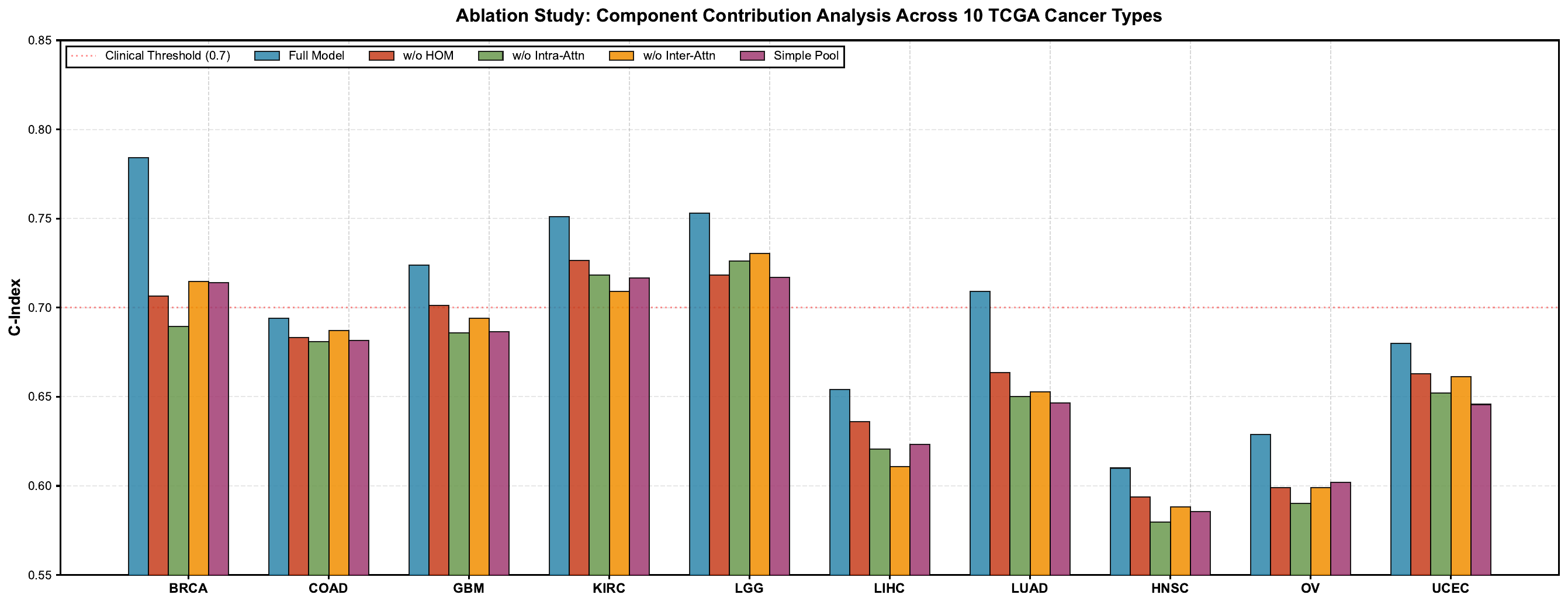}
\caption{\textbf{Ablation study of \pathmog{} components.} Most ablations reduce performance relative to the full model; cohort-specific exceptions are reported from traceable result files.}
\label{fig:ablation_all}
\end{figure}

Beyond the modular graph design, we evaluated which internal components drive performance once pathway modularity is fixed. We therefore tested four ablations across all 10 TCGA cohorts (Figure~\ref{fig:ablation_all}): removing HOM, removing intra-pathway attention, removing inter-pathway attention, and replacing dual attention with simple pooling.

\textbf{Full model remains strongest overall.} The complete model outperformed ablated variants in most cohorts, indicating complementary contributions from modular design, HOM, and hierarchical attention.

\textbf{Intra-pathway attention contributes most among modular components.} Within the ablation benchmark, replacing gene-level attention with mean pooling produced the largest sample-size-weighted drop among non-graph ablations (weighted C-index 0.655 vs.~0.708 for the full model; absolute drop 0.053, relative -7.5\%), with especially large declines in BRCA (-24.0\%), GBM (-8.1\%), and LUAD (-7.4\%). This pattern indicates that weighting genes within each pathway is critical for capturing patient-specific molecular heterogeneity.

\textbf{Component importance is cancer specific.} KIRC and LIHC were most sensitive to inter-pathway attention, LGG depended most on HOM, and UCEC/LUAD showed more balanced dependence across components. These differences support the view that \pathmog{} learns disease-specific multi-omics structure rather than a single universal pattern. Together with the modular-versus-monolithic comparison, these results explain why \pathmog{} achieves stronger benchmark performance.

\subsection{Clinical prognostic analysis}

Having established both benchmark gains and their architectural basis, we next asked whether the \pathmog{} risk score captured information beyond conventional clinical staging. BRCA was used as the main case study because it provided the largest cohort and the clearest downstream treatment-stratification analysis. We fit univariate and multivariate Cox models including age, grade, and TNM variables together with the model-derived risk score.

\begin{table}[t]
\centering
\caption{\textbf{Univariate and multivariate Cox regression for BRCA.} The \pathmog{} risk score remains independently prognostic after adjustment for routine clinical variables.}
\label{tab:cox_analysis}
\resizebox{\linewidth}{!}{%
\begin{tabular}{@{}lccclcc@{}}
\toprule
& \multicolumn{2}{c}{\textbf{Univariate}} & & \multicolumn{2}{c}{\textbf{Multivariate}} \\
\cmidrule(r){2-3} \cmidrule(l){5-6}
\textbf{Variable} & \textbf{HR (95\% CI)} & \textbf{P-value} & & \textbf{HR (95\% CI)} & \textbf{P-value} \\
\midrule
\textbf{Risk score (per SD)} & \textbf{1.62 (1.47--1.78)} & \textbf{<0.001***} & & \textbf{1.57 (1.20--2.07)} & \textbf{0.001**} \\
Age (per SD) & 1.53 (1.30--1.80) & <0.001*** & & 1.20 (0.95--1.53) & 0.130 \\
Grade (G3/G4 vs G1/G2) & 1.66 (1.07--2.58) & 0.023* & & 1.10 (0.53--2.29) & 0.794 \\
T stage (T3/T4 vs T1/T2) & 1.51 (1.03--2.21) & 0.036* & & 0.93 (0.51--1.73) & 0.828 \\
N stage (N1--N3 vs N0) & 1.81 (1.30--2.52) & <0.001*** & & 1.21 (0.77--1.90) & 0.414 \\
M stage (M1 vs M0) & 4.81 (2.88--8.03) & <0.001*** & & 0.77 (0.26--2.26) & 0.628 \\
\bottomrule
\end{tabular}}
\end{table}
The \pathmog{} score was significant in univariate analysis and remained independently prognostic after clinical adjustment (Table~\ref{tab:cox_analysis}). Notably, T stage and N stage were no longer significant after the molecular risk score was added, indicating that \pathmog{} captures prognostic variation not explained by anatomical staging alone. This is the core clinical-statistical claim of the paper: the learned risk score is not a restatement of routine clinicopathological variables.

\subsection{Treatment stratification}

\begin{figure}[t]
\centering
\includegraphics[width=\columnwidth]{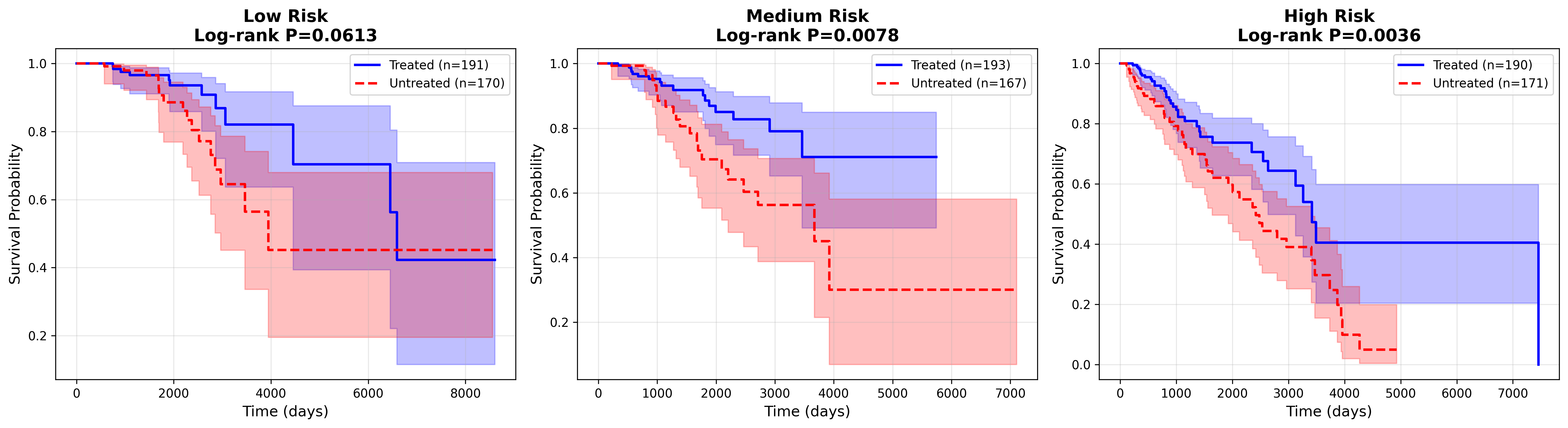}
\caption{\textbf{Treatment response across \pathmog{} risk strata in BRCA.} Benefit from treatment is strongest in the medium- and high-risk groups, whereas the low-risk group shows weaker separation.}
\label{fig:treatment_subgroup}
\end{figure}

We then evaluated whether the \pathmog{} risk score could support treatment-oriented stratification in BRCA. Patients were divided into low-, medium-, and high-risk groups according to the model-derived score, and treatment benefit was compared between treated and untreated patients within each subgroup. Kaplan--Meier subgroup curves are shown in Figure~\ref{fig:treatment_subgroup}.

\begin{figure*}[t]
\centering
\includegraphics[width=\textwidth]{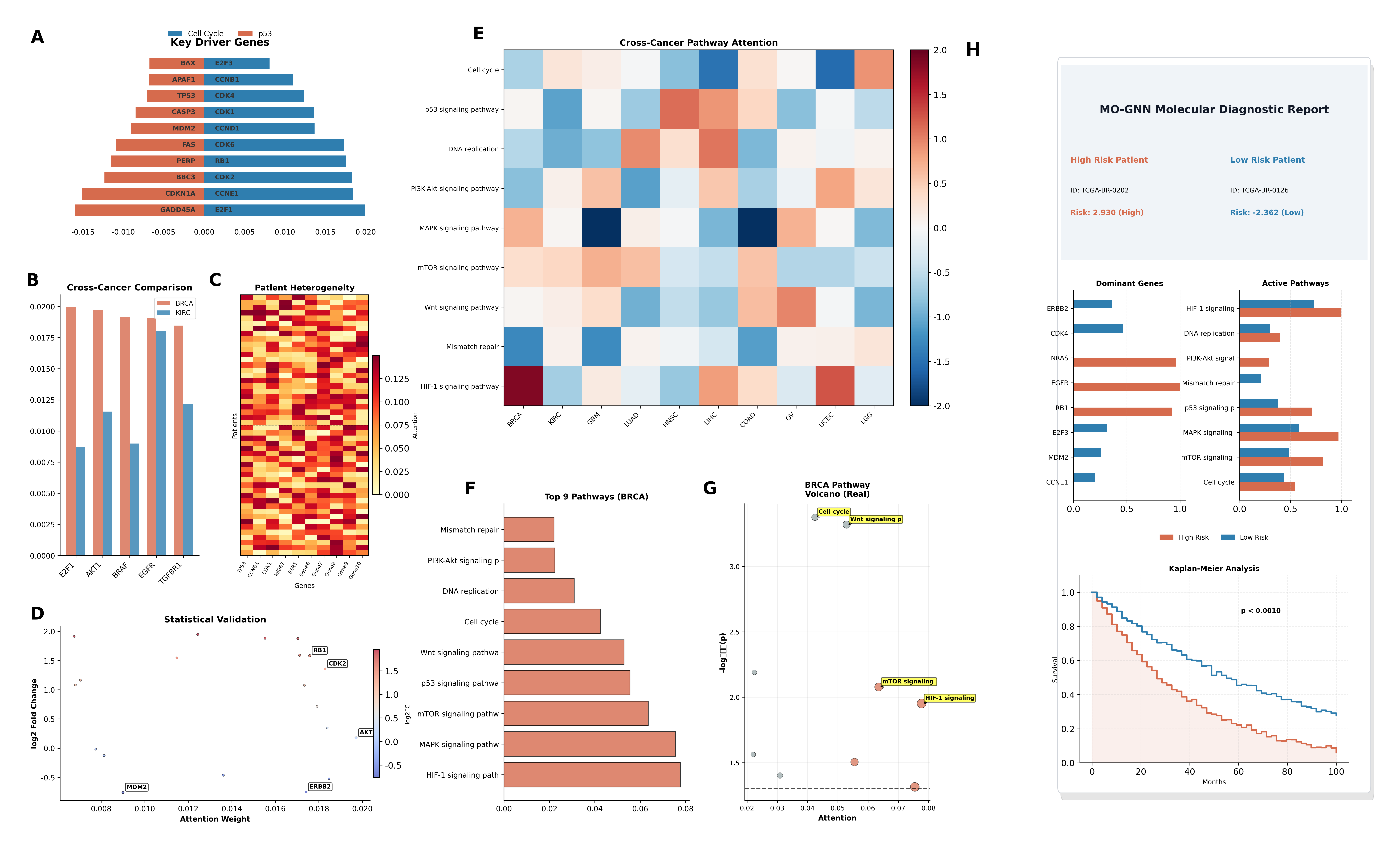}
\caption{\textbf{Multi-level interpretability analysis of \pathmog{}.}
\textbf{(A)} Top key driver genes identified by attention weights in cell cycle and p53 pathways.
\textbf{(B)} Cross-cancer comparison of gene importance between BRCA and KIRC.
\textbf{(C)} Heatmap of patient heterogeneity showing distinct gene expression patterns.
\textbf{(D)} Statistical validation correlating attention weights with differential expression (log2 Fold Change).
\textbf{(E)} Cross-cancer pathway attention heatmap revealing universal and specific pathway dysregulations.
\textbf{(F-G)} Top ranked pathways in BRCA and their statistical significance (volcano plot).
\textbf{(H)} PathMoG Molecular Diagnostic Report illustrating personalized risk assessment, dominant genes, active pathways, and survival prognosis for representative patients.}
\label{fig:combined_interpretability}
\end{figure*}
Treatment benefit was strongest in the medium-risk group ($HR = 0.43$, $p = 0.008$) and remained significant in the high-risk group ($HR = 0.53$, $p = 0.004$), whereas the low-risk group showed only borderline benefit ($HR = 0.49$, $p = 0.061$). This pattern suggests that the \pathmog{} score may help distinguish patients more likely to benefit from systemic therapy from those whose outcomes may not justify equally aggressive intervention.

At the same time, this analysis should be interpreted cautiously. It is retrospective, potentially confounded by treatment-selection bias, and should be treated as hypothesis-generating rather than causal evidence. The value of this section is therefore not to claim treatment guidance as a solved problem, but to show that the learned risk score aligns with clinically meaningful subgroup structure.

\subsection{Biological interpretability}

Finally, \pathmog{} supports comprehensive multi-level interpretability through its dual-level attention mechanism, enabling insights at the gene, pathway, and patient levels (Figure~\ref{fig:combined_interpretability}). To keep the main text concise, we show a compact integrative panel here, while expanded interpretability evidence is provided in Supplementary Section~S11.

\subsubsection{Gene- and pathway-level insights}

By analyzing the learned gene-level attention weights (intra-pathway attention, $\alpha_{g,pw,i}$), we extracted nonlinear feature-importance rankings. Although attention weights should not be treated as a complete explanation on their own \cite{Jain2019}, they provide a useful starting point when combined with differential expression analysis. In the p53 signaling pathway, pro-apoptotic genes such as BAX, APAF1, and TP53 received the highest attention weights, whereas E2F3, CCNE1, and CDK4 emerged as dominant drivers in the Cell Cycle pathway. These findings are consistent with established cancer biology and suggest that \pathmog{} captures survival-relevant dysregulation rather than arbitrary feature salience \cite{Taylor2012, Balko2014}.

Cross-cancer comparison further revealed disease-specific molecular signatures. For example, EGFR and AKT1 received higher attention weights in BRCA than in KIRC, consistent with the stronger contribution of receptor tyrosine kinase signaling to breast cancer progression. Statistical validation within the interpretability panel also showed that high attention-weighted genes, including MDM2, RB1, and ERBB2, were strongly aligned with differential expression patterns, reinforcing the biological relevance of the extracted features \cite{Wade2013}.

At the pathway level, \pathmog{} provides a macroscopic view of cancer progression through pathway attention weights ($\beta_{pw,i}$). Across cohorts, Cell Cycle and p53 signaling appeared as recurrent pan-cancer pathways, whereas BRCA showed stronger emphasis on HIF-1, MAPK, and mTOR signaling. This pattern suggests that the model captures both shared oncogenic programs and tissue-specific pathway dependencies, which is exactly the type of structured readout expected from a pathway-centric design \cite{Marusyk2014, Jiang2022_CAMR}.

\subsubsection{Patient-level insights}

At the patient level, \pathmog{} can consolidate model outputs into an individualized molecular diagnostic profile. For high-risk patients, the model highlights dominant genes, active pathways, and the survival context associated with the predicted score; for lower-risk patients, the same report exposes comparatively quiescent pathway activity and less aggressive molecular signatures. In the original interpretability analysis, this patient-level view was illustrated with a representative report in which elevated EGFR, CDK2, and AKT1 expression co-occurred with strong HIF-1 and PI3K-Akt pathway activity, whereas low-risk patients showed comparatively weaker activation.

This individualized readout is important because it links prediction to actionability. Rather than returning only a scalar risk estimate, \pathmog{} offers a route toward patient-specific explanation: which genes dominate the risk score, which pathways are most active, and how the patient sits within the survival landscape implied by the cohort. That combination of gene-level, pathway-level, and patient-level explanation is the core reason we keep the interpretability module in the main manuscript instead of reducing it to a supplementary-only note.

Detailed interpretability analysis, including cross-cancer pathway heatmaps, cancer-specific pathway rankings, pathway interaction networks, pathway-to-gene decompositions, patient-level diagnostic reports, driver-gene recurrence summaries, and additional gene-level statistical validation, is provided in Supplementary Section~S11.

\section{Conclusion}

\pathmog{} addresses multi-omics survival prediction with a pathway-centric graph design that is explicitly tuned to the $p \gg n$ regime. By replacing a monolithic genome-scale graph with pathway modules, modulating expression with genomic and clinical context, and aggregating molecular evidence through hierarchical attention, the model achieves strong benchmark performance while preserving clinically relevant and biologically interpretable outputs.

Several limitations remain. The present implementation relies mainly on KEGG pathways, uses three omics layers, and currently has external validation from a single major breast cancer cohort. Future work should extend the framework to broader pathway resources, additional molecular modalities, richer external cohorts, and prospective validation settings.

\section{Availability and implementation}

\textbf{Data availability:} TCGA multi-omics and clinical data were obtained from the UCSC Xena Pan-Cancer resource \cite{goldman2020visualizing}. External validation used the METABRIC cohort.

\textbf{Code availability:} Source code for preprocessing, model training, and figure generation is available at \url{https://github.com/wangzoyou/pathmog}.

\section{Supplementary data}

Supplementary data are organized as Sections S1--S12 and are provided separately. They contain the detailed implementation notes, external validation analyses, all-cohort Kaplan--Meier curves, modularity analyses, additional ablation statistics, and extended interpretability figures referenced throughout the main text.

\section*{Author Contributions}
D.W. conceptualized the study, developed the methodology, implemented the models, performed the data analysis, and drafted the manuscript. T.C.L. provided overall research supervision, conceptual guidance, and contributed to manuscript revision. C.P.T., J.X.K., J.X.Z., and M.Y.T. participated in data collection and experimental validation.

\section*{Funding Statement}
This research did not receive any specific grant from funding agencies in the public, commercial, or not-for-profit sectors.

\section*{Key Points}
\begin{itemize}
\item \pathmog{} reframes multi-omics survival prediction as a pathway-centric modular graph problem rather than a genome-scale monolithic graph problem.
\item The HOM module uses mutation, CNV, pathway, and clinical context to modulate gene expression instead of simply concatenating omics features.
\item Across 10 TCGA cancer types, \pathmog{} achieves a sample-size-weighted mean C-index of 0.708 and the highest C-index in all 10 cohorts.
\item The model-derived risk score remains independently prognostic in BRCA after adjustment for routine clinical variables.
\item Extended evidence, including external validation, all-cohort KM curves, design-choice analyses, and richer interpretability panels, is reorganized into a targeted Supplementary file.
\end{itemize}

\bibliographystyle{oup-plain}
\bibliography{refs}

\end{document}